# Decision-Principles to Justify Carnap's Updating Method and to Suggest Corrections of Probability Judgments

Peter P. Wakker
CREED, Dept. of Economics, University of Amsterdam,
Roetersstraat 11, Amsterdam, 1018 WB, The Netherlands

## Abstract

This paper uses decision-theoretic principles to obtain new insights into the assessment and updating of probabilities. First, a new foundation of Bayesianism is given. It does not require infinite atomless uncertainties as did Savage's classical result, and can therefore be applied to any finite Bayesian network. It neither requires linear utility as did de Finetti's classical result, and therefore allows for the empirically and normatively desirable risk aversion. Finally, by identifying and fixing utility in an elementary manner, our result can readily be applied to identify methods of probability updating. Thus, a decision-theoretic foundation is given to the computationally efficient method of inductive reasoning developed by Rudolf Carnap. Finally, recent empirical findings on probability assessments are discussed. It leads to suggestions for correcting biases in probability assessments, and for an alternative to the Dempster-Shafer belief functions that avoids the reduction to degeneracy after multiple updatings.

## 1 INTRODUCTION

This paper examines the quantification of uncertainty through probabilities. In Section 2, we briefly discuss the history of Bayesian decision theory in economics, psychology, statistics, and artificial intelligence, ending with the revived interest generated by Bayesian networks. Section 3 presents a foundation of Bayesianism that is alternative to the classical foundations by de Finetti (1937) and Savage (1954), and that was recently introduced in decision theory. A crucial step in this new foundation is to identify and fix utility (= − loss), so that we can concentrate on the role of probabilities thereafter. Section 4 presents a computationally efficient method for updating probabilities, namely Rudolf Carnap's method of inductive reasoning—I consider this result to be one of the most appealing in the Bayesian field. By means of the technique of Section 3, we give a decision-foundation for Carnap's method. All requirements are expressed exclusively in terms of the empirical primitives, decisions, while at the same time minimizing the role of utilities so that all conditions can readily be interpreted in terms of probabilities. This is the main result of this paper.

Section 5 presents empirical findings, and some recent developments on nonadditive measures, from decision theory. These suggest violations of the convexity conditions that have been traditionally assumed for Dempster-Shafer belief functions and other nonadditive measures. They also suggest nonadditive measures that are alternative to Dempster-Shafer belief functions, and avoid the reduction to degeneracy that hampers the application of the latter in large networks.

## 2 PROBABILITIES AND DECISIONS

The use of probabilities is traditionally justified through Bayesian decision principles, leading to expected utility (de Finetti 1937, Savage 1954). There are, however, several problems for expected utility. Its probabilities must in part be based on subjective assessments, leading to practical and even philosophical problems (Fine 1973). In addition, there are systematic descriptive violations of expected utility. These difficulties have led to non-Bayesian approaches in economics, psychology, and statistics, where nonadditive measures of likelihood play a central role (Machina 1982, Schmeidler 1989, Tversky & Kahneman 1981). In artificial intelligence, a special class of nonadditive measures, the Dempster-Shafer belief functions, became popular (Dempster 1967, McCarthy & Hayes 1969, Shafer 1976). However, coherent manipulations of these measures are problematic.

Whereas non-Bayesian models still flourish in positive sciences such as psychology and economics today, these models, and nonadditive measures, have lost popularity in artificial intelligence, a field oriented towards optimal decision making (Pearl 1988). I hope that the non-Bayesian models still common in statistics today will likewise lose popularity. The revival of Bayesianism also revives some old problems of subjective (or, similarly, logical) probabilities, regarding their ontological status, their measurement, and the decisions derived from them. To the latter we now turn.

Decision theory is sometimes described as probability theory + utility theory, and the use of probabilities in Bayesian networks need not entail a commitment to Bayesian expected utility. Indeed, in an intriguing paper, Machina & Schmeidler (1992) developed a decision theory that uses additive probabilities and Bayes' formula for updating, but does not commit to Bayesian decisions otherwise. I think, however, that this approach cannot serve as a basis for Bayesian statistics or Bayesian networks. A crucial assumption in the latter two, often considered so self-evident that it is assumed implicitly, is the likelihood principle. It implies that, once conditioned on a realization of a random variable (with a specified protocol), the future emerging thereof is not influenced by counterfactual values of that random variable. If this requirement is violated, models become computationally intractable. The requirement is not fulfilled by the model of Machina & Schmeidler, or by any other non-Bayesian evaluation of sequential decisions that does respect Bayes' formula for updating probabilities. Expected utility is, therefore, the only decision model to justify the manipulation of probabilities in Bayesian networks known to me.

## 3 A NEW DECISION FOUNDATION OF BAYESIANISM

As an example to frame our results, we consider the case where treatment decisions have to be taken for a patient with a given set of symptoms. The patient has exactly one disease from a set D of diseases, and it is unknown which. For simplicity of the presentation, we assume that the effects of treatments can be translated into monetary terms. This assumption is made solely for simplicity of the presentation. All results in this paper can easily be adapted to general connected topological spaces, e.g. sets of multidimensional descriptions of health states.

Let $\mathcal{C} \subset \mathbb{R}$, the *outcome* set, be a bounded, closed, and nondegenerate interval that contains zero. A new treatment f yields $f(d) \in \mathcal{C}$ if the patient has disease d. Yielding money means saving money compared to a reference case, e.g. a traditional treatment. Formally, *treatments* (acts in Savage's (1954) terminology) map *diseases* (Savage: states of nature) to outcomes. Subsets of D are *events*. We could consider ($\sigma$-)algebras of events, but for simplicity of the presentation do not do so. For an event A, e.g. the set of lung-diseases, (A:x) denotes the treatment that yields $x if the disease is contained in A, and $0 otherwise. For a treatment f, event A, and $\alpha \in \mathbb{R}$, $\alpha_A f$ denotes the treatment that yields $\alpha$ for all diseases in A, and f(d) for all diseases in $\neg A$. We assume a preference relation $\succcurlyeq$ on the set of treatments, with its symmetric part $\sim$ denoting *equivalence* and its asymmetric part $\succ$ *strict preference*. Preferences and decisions are identified throughout this paper, preferences reflecting binary decisions and decisions consisting of choosing the most preferred treatment.

*(Subjective) expected utility (SEU)* holds if there exists a *(subjective) probability measure* P on D and a *utility function* $U : \mathcal{C} \to \mathbb{R}$, such that preferences are compatible with the *subjective expected utility*

$$f \mapsto \int_D U(f(d))dP$$

of treatments. An event A is *null* if $\alpha_A f \sim \beta_A f$ for all $\alpha,\beta,f$, and it is *nonnull* otherwise. Under SEU with nondegenerate utility, an event is null if and only if its probability is zero.

De Finetti (1937) and Savage (1954) provided classical foundations of Bayesianism. To explain why I will present a different foundation, let us first turn to de Finetti's approach. Assume that $(A:10^6) \sim 300,000$. De Finetti proposed to derive probabilities from betting rates, and would conclude that $P(A) = 3/10$. It is, however, plausible, that $P(A) = 0.5$ and that the equivalence reflects risk aversion. Then, with $\neg$ denoting negation or complement, $(\neg A:10^6) \sim 300,000$ could also be observed, leading to $P(\neg A) = 0.3$, inconsistent with $P(A) = 0.3$. De Finetti based his method on a book making argument, requiring that betting odds cannot be combined into a sure loss. Combining the two equivalences just stated in de Finetti's way would yield $(A:10^6, \neg A:10^6) \sim 600,000$, i.e. the decision maker would readily exchange a million for sure for $600,000 for sure. This violates the book argument, constituting an irrationality according to de Finetti's theory. Still, the original preferences are rational. De Finetti's approach with combinations of preferences is essentially based on the assumption of linear utility, which is not reasonable for large amounts. Restricting attention to small amounts has the drawback that such amounts generate no motivation for taking well-contemplated decisions. Important decisions typically involve large stakes. We will, therefore, propose a foundation of Bayesianism that generalizes de Finetti's argument to nonlinear utility.

Savage's (1954) approach does allow for nonlinear utility, but requires a rich space of uncertain events, which must be infinite and cannot contain atoms (atoms are undividable events). In the medical example considered in this paper, events would have to be considered that specify not only the disease of the patient but also the result of any number of flippings of a coin (Savage 1954, p. 38). Our approach will not require the use of such artificial uncertain events.

Our foundation of Bayesianism extends a result of Köbberling & Wakker (2002), which was obtained for finite disease (state) spaces, to general, possibly infinite, disease spaces. Our foundation uses a method for measuring utility proposed by Wakker & Deneffe (1996). Utility measurement has as yet received little interest in the artificial intelligence community; exceptions are Boutilier (2002) and Poupart et al. (2002). The reason to present this method here is that, by quickly solving the problem of utility measurement, we can abridge the distance between decision theory and probability assessment, and concentrate on the latter. Once our technique established, a decision foundation for Carnap's method of inductive reasoning will readily follow in Section 4.

Consider equivalences

$$\alpha_A f \sim \beta_A g \text{ and} \tag{3.1}$$

$$\gamma_A f \sim \delta_A g \tag{3.2}$$

for a nonnull event A. For $\alpha,\beta$, we write $\alpha\beta \sim^* \gamma\delta$ whenever *there exist* f,g, and a nonnull event A as in (3.1) and (3.2). (3.2) suggests that receiving $\alpha$ instead of $\beta$ can outweigh the same tradeoffs outside of A as receiving $\gamma$ instead of $\delta$, i.e. $\alpha$ instead of $\beta$ is an equally good improvement as $\gamma$ instead of $\delta$. Indeed, substituting expected utility readily shows that

$$\alpha\beta \sim^* \gamma\delta \Rightarrow U(\alpha) - U(\beta) = U(\gamma) - U(\delta). \tag{3.3}$$

The $\sim^*$ relation elicits equalities of utility differences and, as can be seen, the whole utility function, given that utility is an interval scale.

The measurements described should not run into contradictions. A contradiction results, for instance, if we observe, in addition to the two equivalences above, that $\alpha'\beta \sim^* \gamma\delta$, for an $\alpha' > \alpha$. This happens if we find equivalences $\alpha'_B v \sim \beta_B w$ and $\gamma_B v \sim \delta_B w$ for treatments v,w and a nonnull event B, different from f,g,A. Then our utility measurements have run into a contradiction, and expected utility cannot hold. *Tradeoff consistency* requires that contradictory observations as above do not exist. The condition is clearly necessary for expected utility. As we will see, it is also sufficient under some natural conditions, mainly *monotonicity*: $f \succcurlyeq g$ whenever $f(d) \geq g(d)$ for all d, and $\alpha_A f \succ \beta_A f$ for all $\alpha > \beta$, f, and nonnull A.

THEOREM 3.1. Let $\succcurlyeq$ be a nontrivial binary relation defined on the set of treatments, i.e. mappings from a set D to a bounded closed interval $\mathcal{C}$. Then the following two statements are equivalent.

(i) Subjective expected utility holds with a strictly increasing continuous utility function.

(ii) The following conditions hold for $\succcurlyeq$:

    (a) Weak ordering ($\succcurlyeq$ is transitive and complete);

    (b) Monotonicity;

    (c) Supnorm continuity;

    (d) Tradeoff consistency.

□

It is remarkable that the existence of probabilities in SEU follows, free of charge it seems, if we impose consistency of utility measurement through tradeoff consistency. Compared with Savage's (1954) result, the above theorem permits general disease spaces that can be finite or infinite and can have atoms, and does not require a resort to repeated flippings of coins.

*Sigma-additivity* of the probability measure can be ensured by a preference condition requiring that, for each nested sequence of events $A_j$ converging to the empty set and for each $\alpha > 0$, there exists a J such that $(A_j:1) \prec \alpha$ for all $j \geq J$. For infinite sets of diseases it is then desirable to consider a σ-algebra of events, and only measurable treatments.

The technique described above can be used to elicit subjective probabilities without de Finetti's assumption of linear utility. We can in a first stage measure the utility function through revelations of the $\sim^*$ relation, as done in an experiment by Wakker & Deneffe (1996). In a second stage we can derive subjective probabilities from the exchange rate of utility units between different events.

An important reason for developing the utility measurement method described above is that it is, contrary to traditional methods such as the standard gamble, easily adapted to violations of expected utility. With minor adaptations it can, for instance, measure utilities correctly if subjects process probabilities in non-Bayesian manners and do not satisfy additivity. Two experimental papers, Abdellaoui (2000) and Bleichrodt & Pinto (2000), adopted this procedure. They measured decision weights according to the two-stage procedure just described and found that these weights are nonadditive, contrary to the predictions of expected utility.

## 4 UPDATING

Bayesian networks are based on the Bayesian updating of probabilities. To reduce the computational burden, it is desirable that probabilities are taken from *conjugate* parametric families. That is, its members should, after updating, turn into other members of that same parametric family. Well known conjugate families are the beta family and its multinomial extension, the Dirichlet family (Wilks 1962). This section provides a foundation for the use of such families, building on one of the most appealing theorems in the Bayesian field, a theorem by the philosopher Rudolf Carnap (1952, 1980). For an efficient proof and an historical account of Carnap's result, see Zabell (1982). Unfortunately, Carnap's result has not been as well known as it deserves to be.

Carnap expressed his result directly in terms of probabilities. (Subjective) probabilities are, however, not directly observable but either have to be derived from observed decisions, or have to be obtained from introspection and verbal communication, which is a controversial empirical basis for a theory. The empirical status of logical probabilities, Carnap's preferred interpretation, is even less clear. We will derive a foundation directly in terms of the empirical primitive, i.e. decisions.

We assume that D is finite, $D = \{d_1,...,d_s\}$. We have a data set of several patients like the one to be treated now, who were treated in the past. One patient was observed at each time point $i = 1, ..., t$. At time point t, the diseases of all previously observed t patients have become known, leading to the evidence $E = (E^i)_{i=1}^{t}$, with each $E^i \in \{d_1,...,d_s\}$ describing the disease of the ith patient. We consider the same outcome set $\mathcal{C}$ and treatments as in the preceding section, but preferences are conditioned on evidence E, and denoted $\succcurlyeq^E$. Our theory will consider all $\succcurlyeq^E$ for all E and $t \leq T$ for some T with $3 \leq T \leq \infty$. Our conditions will specify how $\succcurlyeq^E$ changes if E changes.

The prior preference relation at time point 0, before any observation is made, is denoted $\succcurlyeq^0$. We will assume that each $\succcurlyeq^E$ satisfies all conditions of the preceding sections, in particular tradeoff consistency, and can be modeled through expected utility.

Our first assumption about updating is that the utility of the outcomes is not affected by the information. Then updating only concerns the probabilities. This assumption is usually made implicitly, but it is nontrivial and can be violated empirically. There can be good reasons for tastes to change over time (Christensen-Szalanski 1984). The decision condition that is necessary and sufficient for identical utilities is that $\sim^{*,E} = \sim^{*,0}$ for each E, and we will assume it. Then utility can be "forgotten" and updating is reduced to variations in probabilities, as is the common case in Bayesian networks. Denote the probabilities conditional on E by $p_i^E$.

Under *Carnap's method of updating*, a positive constant $\lambda$ is chosen. Then each $p_i^E$ is a convex combination of the prior probability $p_i^0 > 0$ and the observed relative frequency $n_i/N$ with weights proportional to $\lambda$ and N, the total number of observations in E. That is,

$$p_i^E = \frac{\lambda p_i^0 + N\frac{n_i}{N}}{\lambda + N}. \quad (4.1)$$

The parameter $\lambda$ indicates the strength of evidence of the prior information, relative to the strength of evidence of the observations of the other patients. $\lambda$ will be bigger as the prior evidence is more reliable, and smaller as the other patients are more closely related to the one now under consideration. We can interpret $\lambda$ as equivalent to a hypothetical sample size. Indeed, (4.1) is the relative frequency of a sample consisting of $\lambda$ observations with relative frequency $p_i^0$ and N observations with relative frequency $n_i/N$.

Carnap's method is appealing, but may seem ad hoc if posited without foundation. People are reluctant to choose subjective parameters such as $p_i^0$ and $\lambda$. There are even philosophical objections against the use of such subjective quantities (Fine 1973). Questions can be raised: Could not this formula, while plausible at face value, lead us astray and yield implausible values? Why are the operations chosen as they are? Why not take a geometrical mean, instead of an arithmetical mean, of $p_i^0$ and $n_i/N$? How can we convince ourselves of the appropriateness of the operations chosen?

Carnap gave an answer to the preceding questions, taking probabilities as primitive. As explained, we shall reformulate Carnap's answers directly in terms of decisions, so as to obtain a clear normative and empirical status of Carnap's induction method.

We distinguish sharply between our conditions and the interpretations thereof. Our conditions should provide a sound and uncontroversial empirical basis and, therefore, will only concern the $\succcurlyeq^E$ relations, treatment decisions for single patients given fixed evidence. The interpretations of the conditions serve to give intuitive insights into the conditions, and here we can use whatever we find helpful. Here we will freely refer to probabilities and to hypothetical decisions concerning sequences of patients. Because our conditions only concern conditional probabilities given evidence E, and never joint probabilities concerning sequences of patients or all of them, our formal treatment need not state Bayes' conditioning formula explicitly. The interpretation and reasonableness of exchangeability, hereafter, will, however, be essentially based on Bayes' conditioning. Figure 1 illustrates the conditions discussed next.

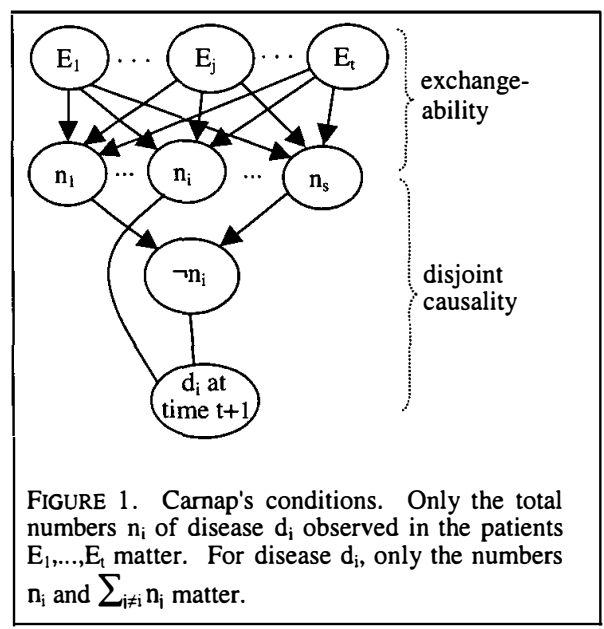

FIGURE 1. Carnap's conditions. Only the total numbers $n_i$ of disease $d_i$ observed in the patients $E_1,...,E_t$ matter. For disease $d_i$, only the numbers $n_i$ and $\sum_{j \neq i} n_j$ matter.

Our first condition is *positive relatedness of the observations*:

$$(d_i:1) \sim^E x \Rightarrow (d_i:1) \succ^{(E,d_i)} x. \qquad (4.2)$$

Any new observation of disease $d_i$, added to the evidence E already available, makes it more attractive to gain \$1 if the patient now to be treated will have disease $d_i$. We interpret (4.2) as meaning that $d_i$ is considered more likely after an extra observation $d_i$. This condition will be violated in specific cases, e.g. in sampling without replacement, where observations are related negatively. Our main theorem, in fact, goes through with minor adaptations if observations are negatively related, and we assume Eq. 4.2 mostly for simplicity of the presentation. The essential mathematical point is that the observations are not independent, i.e. the connections with $d_i$ at time t+1 in Figure 1 should be maintained. Let us assume henceforth that positive relatedness is appropriate in our empirical domain.

For the interpretation of the next condition, we start by defining the joint probability of a sequence of observations $(d_{i_1},...,d_{i_t})$ through the product of conditional probabilities, $p_{i_1}^0 \times p_{i_2}^{E'_1} \times \cdots \times p_{i_t}^{E'_{t-1}}$, where each $E'_j$ consists of the first j observations $(d_{i_1},...,d_{i_j})$. The product of probabilities just defined requires, for a proper interpretation, Bayes' conditioning. Because these probabilities play a role only in the interpretation of the following condition, and not in the formal theory, our formal framework need not specify Bayes' conditioning.

For $E = (E_1,...,E_t)$, $(E,d_i)$ denotes $(E_1,...,E_t,E_{t+1})$ with $E_{t+1} = d_i$, $(E,d_i,d_j)$ denotes $(E_1,...,E_t,E_{t+1},E_{t+2})$ where furthermore $E_{t+2} = d_j$, etc. Our next decision principle characterizes exchangeability. Consider equivalences

$$(d_j:1) \sim^{(E,d_i)} x \text{ and } (d_i:x) \sim^E y. \qquad (4.3)$$

We start from evidence E. The first equivalence states that, if an extra observation $d_i$ is made, then gaining \$1 conditional on $d_j$ is equivalent to gaining \$x for sure. From the perspective of evidence E preceding any extra observation $d_i$, this equivalence can be reinterpreted as an equivalence between:

* gaining \$1 if first $d_i$ is observed and then $d_j$, and gaining nothing if the next observation is not $d_i$;

* gaining \$x always if first $d_i$ is observed, and gaining nothing if the next observation is not $d_i$.

Whereas our formal model, and conditions, only consider $\succeq^E$, decisions regarding one patient, being the next patient, the reinterpreted equivalence, serving only to clarify, refers to hypothetical decisions based on two successive patients. The second equivalence in Eq. 4.3 simply states that gaining \$y for sure is equivalent to gaining \$x if $d_i$ is observed, all given E. Combining the second and reinterpreted first equivalences suggests that receiving \$y for sure is equivalent to receiving \$1 if first $d_i$ is observed and then $d_j$.

Next consider the two equivalences

$$(d_i:1) \sim^{(E,d_j)} x' \text{ and } (d_j:x) \sim^E y'. \qquad (4.4)$$

They can be interpreted similarly as Eq. 4.3, suggesting that receiving \$y' for sure is equivalent to receiving \$1 if first $d_j$ is observed and then $d_i$. *Exchangeability* requires that y in (4.3) and y' in (4.4) are identical. This condition suggests that first observing $d_i$ and then $d_j$ is as probable as first observing $d_j$ and then $d_i$. By repeated application, the condition implies that the probability of a sequence of observations depends only on the number of observations of each disease in the sequence, and not on the order in which these observations were made. The condition, therefore, agrees with the exchangeability condition common in Bayesian statistics. It implies that the sampling of patients could as well have been done in an unordered manner. Although the condition can be violated if causes change over time, let us assume that the condition is appropriate in our empirical domain.

The decision conditions considered so far, positive relatedness and exchangeability, are directly related to assumptions common in statistics. The following condition is less common, and is characteristic of Carnap's result. It amounts to the assumption that there are no common causes for different diseases, i.e. disease $d_1$ is equally probable after observing $d_2$ as after observing $d_3$. Formally, *disjoint causality* means that, for all E and distinct i,j,k,

$$(d_i:1) \sim^{(E,d_j)} x \Rightarrow (d_i:1) \sim^{(E,d_k)} x. \qquad (4.5)$$

Under exchangeability, it only matters for the conditional probability of a disease $d_i$ how many times we observed $d_i$ and $\neg d_i$, and not whether among the $\neg d_i$ observations there were more of one disease or of another. For s = 2, the condition is vacuously satisfied and Theorem 4.1 will not apply to s=2.

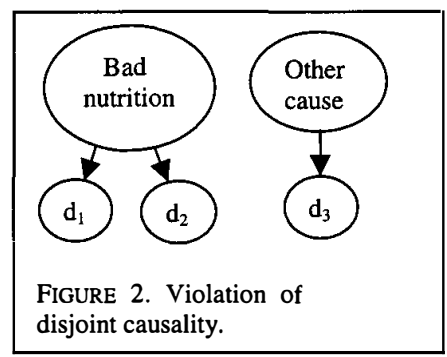

FIGURE 2. Violation of disjoint causality.

Disjoint causality will not be appropriate in many situations. The causal configuration in Figure 2 is a typical case. If there is a common cause for diseases $d_1$ and $d_2$, then an observation of $d_2$ can, more than an observation $d_3$, be indicative of $d_1$ in the future, and disjoint causality is violated. Disjoint causality is plausible if all diseases are generated by independently operating families of bacterias, to whom patients are delivered on a first-come-first-serve basis.

Let us assume that disjoint causality is reasonable in our empirical domain. One disease is not closer to a second than to a third, and we take diseases as nominal. There is no structure on the set of diseases that we can use in our inferences.

The following theorem may come as a surprise. The first issue raised above, the reasonableness and empirical meaning of Carnap's inductive method, was hard to assess. The meaning of its quantities was not directly related to observable primitives, and the method seemed to be ad hoc. The second issue raised was the reasonableness of a number of conditions regarding the structure of our reasoning and our decisions, depicted in Figure 1. They were all stated directly in terms of decisions, and their empirical status was clear. We assumed that these conditions are appropriate in our particular empirical domain. As it turns out, the two issues raised are equivalent, and Carnap's inductive method is appropriate if and only if the conditions discussed are appropriate.

THEOREM 4.1. Assume $s \geq 3$. The following two statements are equivalent.

(i) Carnap's inductive method (Eq. 4.1) holds, where $\succcurlyeq^E$ maximizes subjective expected utility with respect to the $p_i^E > 0$ and a strictly increasing continuous utility function U.

(ii) Each $\succcurlyeq^E$ satisfies the conditions of Statement (ii) in Theorem 3.1, with all diseases nonnull. Further, the following conditions are satisfied.

(a) $\sim^{*,E} = \sim^{*,0}$ for each E;

(b) Positive relatedness of the observations;

(c) Exchangeability;

(d) Disjoint causality of the diseases.

The probabilities and the number $\lambda$ in Carnap's method are uniquely determined through $\succcurlyeq^E$.

□

An intelligent client without training in subjective probabilities can more easily be convinced through Statement (ii) than through Statement (i). In Theorem 4.1, Carnap's method, with its subjective and philosophically controversial quantities, follows from a number of empirically meaningful claims as a logical necessity, without further resort to philosophical arguments. Decision foundations for other updating systems, such as for sets of probabilities (Boutilier, Friedman, & Halpern 2002), remains a topic for future research.

Carnap's formula has many attractive features. It shows how to integrate subject-matter information ($p_i^0$) and statistical information ($n_i/N$). The subject-matter information $p_i^0$ can be interpreted as being equally informative as $\lambda$ observations. If many observations become available, the role of the problematic subjective parameters can be ignored and the observed relative frequency decides. The formula can also be used for combining two pieces of evidence instead of updating (Halpern & Fagin 1992), where the weight of each piece is calibrated through a hypothetical number of observations that are considered equally informative.

## 5 EMPIRICAL FINDINGS

This section discusses empirical findings regarding subjective probabilities derived from decisions. Section 3 suggested a simple procedure for measuring utility, which can subsequently be used to measure subjective probabilities. It is not guaranteed empirically that the quantities measured will behave like probabilities—Abdellaoui (2000) and Bleichrodt & Pinto (2000) indeed found deviations. We shall, therefore, use the symbol W and not P to denote the results of these measurements. This section is informal and we will, therefore, not specify the nonexpected utility theories underlying our claims. It suffices to note that treatments (A:x) are evaluated through W(A)U(x) where W may be a nonadditive measure, e.g. a Dempster-Shafer belief function, and where, for convenience, U(0)=0. Then W(A)=U(x)/U(1) can be derived from an observed preference (A:1)~x. For simplicity, we only consider nonnegative outcomes.

A first point of concern under violations of Bayesianism may be that the measurements of utility of Section 3 need no longer be valid. Fortunately, these measurements are easily adapted to be robust to the violations of expected utility mostly studied today (Wakker & Deneffe 1996). We will, therefore, concentrate on the deviations from additivity comprised in W.

Tversky & Fox (1995) conjectured a two-stage model W = w ○ φ, where φ is a direct psychological judgment of probability, and w turns judgments of probability into decision weights. The component w can be measured from the special cases where objective probabilities of events are known, so that φ is the identity. The shape of w that is mostly conjectured by economists and by the artificial intelligence community is convexity. This conjecture is based on theoretical arguments, because convexity is related to conservative and pessimistic risk attitudes and to the existence of equilibria. Convexity of w enhances convexity of W, implying for instance that

$$W(A \cup B) \geq W(A) + W(B) \text{ (superadditivity)} \quad (5.1)$$

for disjoint events A,B, a condition satisfied by the Dempster-Shafer belief functions.

Psychologists are empirically oriented and, rather than conjecture shapes of W based on theoretical or normative arguments, they carried out measurements to see what w and W look like empirically. The prevailing shape for w that was found is different from what was conjectured elsewhere, and is depicted in Figure 3 (Abdellaoui 2000, Bleichrodt & Pinto 2000, Tversky & Fox 1995).

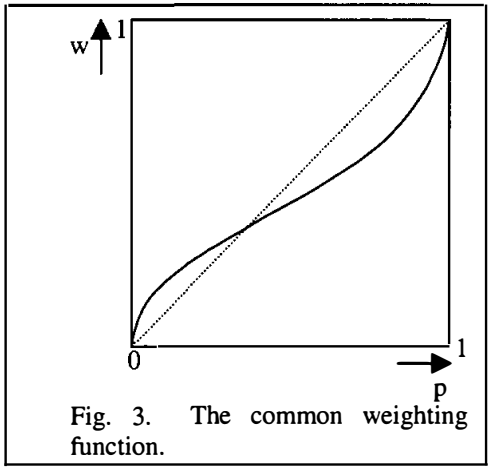

Fig. 3. The common weighting function.

Small probabilities are overweighted, contrary to convexity as mostly assumed in the literature. For probabilities p, q that are small or moderate, we have

$$w(p+q) \leq w(p) + w(q) \text{ (subadditivity)}. \qquad (5.2)$$

The w component of W enhances subadditivity of W,

$$W(A \cup B) \leq W(A) + W(B) \qquad (5.3)$$

for disjoint events A,B, contrary to the common assumptions about belief functions etc. as in Eq. 5.1.

A natural idea when inferring probabilities from decisions, is to first derive W and w directly from decisions, then to "correct" W for w, that is, to take $\varphi = w^{inv}W$ as the judgment of the expert. A question is whether all nonlinearity in W is due to w, so that $w^{inv}W$ can play the role of a Bayesian belief. The empirical findings so far suggest that this is not so (Tversky & Fox 1995). Much, even most, of the deviation from Bayesianism lies in the estimation of unknown probabilities $\varphi$. The domain of $\varphi$, as of W, is the set of events and not the unit interval and, therefore, no graph of $\varphi$ or W can be drawn. Still, $\varphi$'s and W's deviations from linearity are of the same nature as those of w depicted in Figure 3. Tversky & Wakker (1995) five a formal definition, called *bounded subadditivity*.

For those who do not accept the Bayesian paradigm as normative, the developments just described lead to suggestions for nonadditive measures alternative to the Dempster-Shafer belief functions. A difficulty with the latter is that after iterated updatings they lead to degenerate assignments, with the belief of most events converging to zero, or to one if the, dual, plausibility functions are used. Such degenerate quantifications resulting from the absence of information can no more be updated flexibly if valuable information arrives after. The nonadditive measures with properties similar to those depicted in Figure 3, tend to some interior value, such as 0.5 in the case of symmetry, and not to zero or one. Such quantifications are flexibly updated when valuable information starts arriving at some stage, thus avoid one of the major problems of belief functions.

The shape of Figure 3, and its analogue for $\varphi$ and W, suggest a lack of sensitivity towards varying degrees of uncertainty because the function is too shallow in the middle. This constitutes a cognitive reason for deviating from Bayesianism, instead of the motivational reason suggested by the pessimism that has been assumed in economics and artificial intelligence so far. The shape in Figure 3 seems to better reflect absence of information than convexity, both psychologically and computationally. Also linguistically, absence of information is expressed by "fifty-fifty" rather than by degenerate assessments (Fischhoff & Bruine de Bruin 1999). Although the phenomena of subadditivity and Figure 3 did not originate from conceptual considerations, but from inspections of data, these phenomena do suggest new concepts such as degrees of sensitivity towards information rather than degrees of conservativeness or pessimism.

For those who, like the author, consider the Bayesian paradigm as normative, the developments sketched in this section suggest how to correct expert judgments when elicited from (hypothetical) decisions, so as to bring them closer to the Bayesian approach.

To end this discussion of empirical findings, I briefly mention support theory for direct judgments of probabilities that are not revealed through decisions. The psychologist Amos Tversky developed this theory in the last years of his life (Tversky & Koehler 1994). A typical finding of support theory is, again, subadditivity, which increases with refinements. That is, for disjoint events $A_j$,

$$\varphi(A_1) + \cdots + \varphi(A_n) - \varphi(A_1 \cup \cdots \cup A_n) \qquad (5.4)$$

increases as n increases.

UAI 2002                                    WAKKER                                                551Boutilier, Craig (2002), "A POMDP Formulation of Preference Elicitation Problems," Dept. of Cumputer Science, University of Toronto, Toronto, Canada.

Boutilier, Craig, Nir Friedman, & Joseph Y. Halpern (2002), "Belief Revision with Unreliable Observations," Dept. of Cumputer Science, University of Toronto, Toronto, Canada.

Carnap, Rudolf (1952), *The Continuum of Inductive Methods*." University Press, Chicago.

Carnap, Rudolf (1980), "A Basic System of Inductive Logic, Part II." In Richard C. Jeffrey (Ed.), *Studies in Inductive Logic and Probability*, Vol. II, 7–155, University of California Press, Berkeley.

Christensen-Szalanski, Jay J.J. (1984), "Discount Functions and the Measurement of Patients' Values; Woman's Decisions During Childbirth," *Medical Decision Making* 4, 47–58.

Dempster, Arthur P. (1967), "Upper and Lower Probabilities Induced by a Multivalued Mapping," *Annals of Mathematical Statistics* 38, 325–339.

Fine, Terrence L. (1973), "*Theories of Probability*." Academic Press, New York.

de Finetti, Bruno (1937), "La Prévision: Ses Lois Logiques, ses Sources Subjectives," *Annales de l'Institut Henri Poincaré* 7, 1–68. Translated into English by Henry E. Kyburg Jr., "Foresight: Its Logical Laws, its Subjective Sources," in Henry E. Kyburg Jr. & Howard E. Smokler (1964, Eds.), *Studies in Subjective Probability*, 53–118, Wiley, New York; 2nd edition 1980, Krieger, New York.

Fischhoff, Baruch & Wändi Bruine de Bruin (1999), "Fifty-Fifty = 50%?," *Journal of Behavioral Decision Making* 12, 149–163.

Halpern, Joseph Y. & Ronald Fagin (1992), "Two Views of Belief: Belief as Generalized Probability and Belief as Evidence," *Artificial Intelligence* 54, 275–317.

Köbberling, Veronika & Peter P. Wakker (2002), "A Tool for Qualitatively Testing, Quantitatively Measuring, and Normatively Justifying Expected Utility," METEOR, Maastricht University, The Netherlands.

Machina, Mark J. & David Schmeidler (1992), "A More Robust Definition of Subjective Probability," *Econometrica* 60, 745–780.

Machina, Mark J. (1982), " 'Expected Utility' Analysis without the Independence Axiom," *Econometrica* 50, 277–323.

McCarthy, J. & P. Hayes (1969), "Some Philosophical Problems from the Standpoint of Artificial Intelligence." In B. Meltzer & D. Michie (Eds.), Machine Intelligence Vol. 4, 463–502, Edinburgh University Press, Edinburgh, U.K.

Poupart, Pascal, Craig Boutilier, Relu Patrascu, & Dale Schuurmans (2002), "Piecewise Linear Value Function Approximation for Factored MDPs," Dept. of Cumputer Science, University of Toronto, Toronto, Canada.

Pearl, Judea (1988), "*Probabilistic Reasoning in Intelligent Systems: Networks of Plausible Inference*." Morgan Kaufmann, San Mateo CA.

Savage, Leonard J. (1954), "*The Foundations of Statistics*." Wiley, New York. (Second edition 1972, Dover, New York.)

Schmeidler, David (1989), "Subjective Probability and Expected Utility without Additivity," *Econometrica* 57, 571–587.

Shafer, Glenn (1976), "*A Mathematical Theory of Evidence*." Princeton University Press, Princeton NJ.

Tversky, Amos & Craig R. Fox (1995), "Weighing Risk and Uncertainty," *Psychological Review* 102, 269–283.

Tversky, Amos & Daniel Kahneman (1981), "The Framing of Decisions and the Psychology of Choice," *Science* 211, 453–458.

Tversky, Amos & Derek J. Koehler (1994), "Support Theory: A Nonextensional Representation of Subjective Probability," *Psychological Review* 101, 547–567.

Tversky, Amos & Peter P. Wakker (1995), "Risk Attitudes and Decision Weights," *Econometrica* 63, 1255–1280.

Wakker, Peter P. & Daniel Deneffe (1996), "Eliciting von Neumann-Morgenstern Utilities when Probabilities Are Distorted or Unknown," *Management Science* 42, 1131–1150.

Wilks, Samuel S. (1962), "*Mathematical Statistics*." Wiley, New York

Zabell, Sandy L. (1982), "W.E. Johnson's "Sufficientness" Postulate," *The Annals of Statistics* 10, 1091–1099.